\documentclass[final,1p,times,twocolumn]{elsarticle}
\usepackage{amssymb}
\usepackage{amsmath}
\usepackage{amsmath,amssymb,amsfonts}
\usepackage{algorithmic}
\usepackage{graphicx}
\usepackage{algorithm,algorithmic}

\usepackage{textcomp}
\usepackage{multirow}
\usepackage{graphicx}
\usepackage{hyperref}
\usepackage[table,xcdraw]{xcolor}
\definecolor{turquoise}{RGB}{61,130,125}
\definecolor{red}{RGB}{195,85,58}

\journal{Computers in Biology and Medicine}

\makeatletter
\def\ps@pprintTitle{%
  \let\@oddhead\@empty
  \let\@evenhead\@empty
  \def\@oddfoot{}%
  \let\@evenfoot\@oddfoot}
\makeatother

\begin{document}

\begin{frontmatter}



\title{Personalized Weight Loss Management through \\Wearable Devices and Artificial Intelligence}

\author[UAM]{Sergio Romero-Tapiador\corref{cor1}}
% \author[UAM]{Sergio Romero-Tapiador}
\author[UAM]{Ruben Tolosana} 
\author[UAM, LPGC]{Aythami Morales} 
\author[IMDEA1]{Blanca Lacruz-Pleguezuelos} 
\author[UAM]{Sofia Bosch Pastor} 
\author[IMDEA1]{Laura Judith Marcos-Zambrano} 
\author[IMDEA2]{Guadalupe X. Bazán} 
\author[IMDEA2]{Gala Freixer} 
\author[UAM]{Ruben Vera-Rodriguez} 
\author[UAM]{Julian Fierrez} 
\author[UAM]{Javier Ortega-Garcia} 
\author[IMDEA2]{Isabel Espinosa-Salinas} 
\author[IMDEA1]{Enrique Carrillo de Santa Pau}

\affiliation[UAM]{organization={BiometricsAI, Universidad Autonoma de Madrid},%Department and Organizati
            city={Madrid},
            postcode={28049}, 
            country={Spain}}

\affiliation[LPGC]{Department of Mathematics, Universidad de Las Palmas de Gran Canaria, 35001, Spain}

\affiliation[IMDEA1]{organization={Computational Biology Group, Precision Nutrition and Cancer Research Program, IMDEA Food Institute},
            addressline={CEI UAM+CSIC}, 
            city={Madrid},
            postcode={28049}, 
            country={Spain}}          

\affiliation[IMDEA2]{organization={GENYAL Platform on Nutrition and Health, IMDEA Food Institute},
            addressline={CEI UAM+CSIC}, 
            city={Madrid},
            postcode={28049}, 
            country={Spain}}    

\cortext[cor1]{Corresponding author. E-mail: \href{mailto:sergio.romero@uam.es}{sergio.romero@uam.es}}

%% Abstract
\begin{abstract}
 Early detection of chronic and non-communicable diseases (NCDs) is crucial for effective treatment during the initial stages. This study explores the application of wearable devices and artificial intelligence (AI) to predict weight loss changes in overweight and obese individuals. To the best of our knowledge, this is the first study to leverage wearable-derived physiological and behavioral features for weight loss prediction using AI. Using wearable data from a 1-month trial involving around 100 subjects from the AI4FoodDB database, including biomarkers, vital signs, and behavioral data, we identify key differences between those achieving weight loss ($\geq$ 2\% of their initial weight) and those who do not. Feature selection techniques highlight key predictors, including heart rate variability, sleep duration, and physical activity patterns, underscoring the interplay between physiological and behavioral factors. Classification algorithms reveal promising results, with the Gradient Boosting classifier achieving 84.44\% area under the curve (AUC). The integration of multiple data sources (e.g., vital signs, physical and sleep activity, etc.) enhances performance, suggesting the potential of wearable devices and AI in personalized healthcare. 
\end{abstract}


% \begin{highlights}
% \end{highlights}

%% Keywords
\begin{keyword}
    Wearable Devices\sep Personalized Healthcare\sep Weight Loss\sep Nutrition\sep Artificial Intelligence\sep AI4FoodDB

\end{keyword}

\end{frontmatter}

\section{Introduction}\label{sec:intro}

 Non-communicable diseases (NCDs) represent 74\% of all deaths globally, and traditional healthcare methods fail to curb the increase of NCDs in current society \cite{WHONCDs}. Most NCDs are influenced by highly modifiable factors, such as nutrition, physical activity (PA), sleep activity (SA), and medication, as well as unmodifiable factors like age, gender, genetics, and socio-cultural influences \cite{michel2019nutrition}. In 2022, according to the World Health Organization (WHO), 81\% of adolescents and 27.5\% of adults fail to meet the minimum PA guidelines, which recommend at least 150 minutes of moderate-intensity PA or 75 minutes of vigorous-intensity PA per week \cite{world2022global}. The COVID-19 pandemic has further aggravated the reduction in PA, contributing to increased rates of obesity and related NCDs, such as heart diseases and type 2 diabetes \cite{natalucci2023effectiveness}. To put some figures, in 2022 approximately 2.5 billion adults were overweight, and over 890 million suffered from obesity \cite{oms_obesity}.

\begin{figure}
    \centering
    \includegraphics[width=0.98\linewidth]{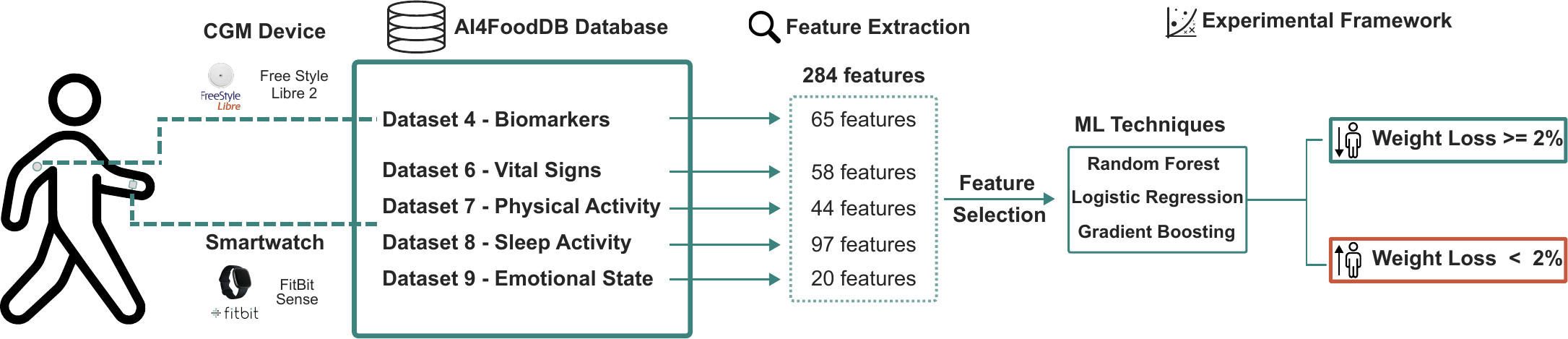}
    \caption{Description of the application scenario and machine learning (ML) methods proposed in this study for the prediction of weight loss. The experimental framework integrates data acquired from continous glucose monitor (CGM) devices and smartwatches. In particular, we consider the AI4FoodDB database \cite{romero2023ai4fooddb}, considering five datasets covering physiological and lifestyle data, including Biomarkers (Dataset 4), Vital Signs (Dataset 6), Physical Activity (Dataset 7), Sleep Activity (Dataset 8), and Emotional State (Dataset 9). From these datasets, in the present article we propose the extraction of 284 total features that undergo feature selection before being evaluated using ML techniques. The proposed ML models ultimately distinguish between subjects who lost $\geq$ 2\% of their initial weight and those who did not.}
    \label{fig:graphabstr}
\end{figure}

Promoting a healthy lifestyle and behaviors can significantly reduce potential risk factors associated with NCDs. However, the increasing global population, healthcare costs, and limited access to individualized medical attention present major challenges to effective healthcare delivery \cite{subhan2023ai}. In this context, personalized healthcare, which integrates digital and non-digital data, is becoming very important in order to alleviate the financial and personnel burdens on the healthcare system. Advances in digital technologies, such as the internet of things (IoT), have led to the development of new sensors and wearable devices that enable human-computer interactions (HCI). Initially, these devices focused on recording exercise and heart rate (HR), but current sensors also measure various physiological parameters not only for detecting any pathological issue but also for monitoring the lifestyle and behavior of individuals \cite{CHEN2018396}. These technologies facilitate the creation of digital twins, which allow for continuous monitoring and rapid communication between patients and healthcare professionals, consequently, reducing the healthcare system workloads \cite{williams2023wearable, KALRA2024108917}.

Artificial intelligence (AI), particularly approaches based on machine learning (ML) and deep learning (DL), are essential for processing the vast amounts of continuous data generated by wearable devices \cite{neri1}. These techniques can detect minor physiological changes from baseline values, enabling the development of personalized models that outperform traditional population-level models \cite{MAHMOUD2023107295}.

Furthermore, the use of wearable devices and smartphones has grown exponentially over the past decade, generating significant social and economic impacts \cite{xie2020electronic}. These devices facilitate rapid data transfer, automatic self-monitoring, and the generation of digital health information that promotes behavioral changes and health improvements. Therefore, society is increasingly adopting self-monitoring practices for tracking lifestyle, PA, SA,  nutrition, and overall health status \cite{ferguson2022effectiveness, PATLARAKBULUT2020101824}. However, several challenges remain in leveraging these technologies effectively. Different companies develop sensors that track similar physiological parameters, leading to a need for standardization in both sensor specifications and data processing. Additionally, data accessibility and privacy face several challenges \cite{melzi2022overview}. First, most consumer-based wearable devices are designed for emerging adults and are often unaffordable for low-income and minority populations. Also, there is skepticism about the use of invasive devices \cite{zinzuwadia2022wearable}.

Digital transformation promises to enhance the patient-doctor relationship, even in the absence of apparent disease. While time is still needed to develop the precise technology that integrates all acquired data from wearable devices, progress has been made in promoting health through wearable devices that encourage PA and mental wellness by sending notifications and reports about self-monitoring \cite{ferguson2022effectiveness, zahrt2023effects}. 

The main problem addressed in the present study is whether it is possible to predict weight loss or gain solely based on data acquired from wearable devices (smartwatches and glucose sensors).  Fig.~\ref{fig:graphabstr} provides a graphical representation of the application scenario and ML methods proposed in the present study. By analyzing extensive data collected from these devices, and using ML techniques through the proposal of a novel set of 284 features related to biomarkers, vital signs, PA, SA, and emotional state, this study aims to identify weight change patterns and predictors, enhancing weight loss management and improving the effectiveness of self-monitoring.

The rest of this paper is organized as follows. State-of-the-art studies related to the use of wearable devices and AI in predicting health outcomes are presented in Sec. \ref{sec:relatedworks}. Sec. \ref{sec:datasets} describes the AIFoodDB database, considered in the experimental framework of the study. Sec. \ref{sec:methods} explains the proposed methods to predict weight loss whereas Sec. \ref{sec:experiments} describes the experimental results. Finally, conclusions and future studies are drawn up in Sec. \ref{sec:conclusion}.

\section{Related Works}\label{sec:relatedworks}

 Numerous studies have demonstrated the benefits of wearable technologies and adherence to digital self-monitoring in weight loss interventions \cite{berry2021does}. For instance, Mao \textit{et al.} proved that PA interventions effectively promote weight loss in overweight and obese individuals using mobile phones as trackers \cite{mao2017mobile}. In \cite{fazzino2017change}, they examined weight loss and maintenance by analyzing PA via accelerometer and self-report in breast cancer survivors, highlighting the importance of using wearable devices. However, most of these studies considered only a few general features derived from wearables to analyze behavioral changes related to weight loss \cite{hutchesson2018targeted}.

Contrarily, many other studies have considered newly implemented features extracted from wearable devices and they employ ML techniques to predict clinical outcomes using clinical data (i.e., anthropometric measurements and laboratory values) and digital data (e.g., wearable devices and digital health) \cite{yen2019effectiveness}. Table \ref{tab:sota} presents some key studies found in the literature on topics related to the present study (i.e., PA and SA, emotional state, etc.).

% However, one of the main disadvantages of using these devices is the lack of standards that validate their quality, often resulting in measurement errors. For instance, Dooley \textit{et al.} \cite{dooley2017estimating} found significant differences in heart rate-related features across different commercial smartwatches, linking these discrepancies to factors such as design, materials, and software.

% \usepackage{graphicx}
\begin{table}[]
\caption{Overview of key state-of-the-art studies utilizing wearable devices for the prediction of health outcomes. For each study, we list authors, main goal, comments, and results achieved. RMSE = Root Mean Square Error.}
\label{tab:sota}
\resizebox{\textwidth}{!}{%
\begin{tabular}{cccc}
\hline
\textbf{Authors} & \textbf{Study Goal} & \textbf{Comments} & \textbf{Results Achieved} \\ \hline
\begin{tabular}[c]{@{}c@{}}Beattie \textit{et al.} \\ \cite{beattie2017estimation} (2017)\end{tabular} & \begin{tabular}[c]{@{}c@{}}Sleep stage \\ classification\end{tabular} & \begin{tabular}[c]{@{}c@{}}Automated classifiers using\\ 180-wrist-worn-device features.\end{tabular} & Cohen's Kappa = 0.52\\
\multicolumn{1}{l}{} & \multicolumn{1}{l}{} & \multicolumn{1}{l}{} &  \\
\begin{tabular}[c]{@{}c@{}}Zhang \textit{et al.}\\ \cite{zhang2018sleep} (2018)\end{tabular} & \begin{tabular}[c]{@{}c@{}}Sleep stage \\ classification\end{tabular} & \begin{tabular}[c]{@{}c@{}}Sleep-related features collected \\ from wearable devices to feed RNNs.\end{tabular} & \begin{tabular}[c]{@{}c@{}}Precision = 66.6\%\\ Recall = 67.7\%\\ F1 Score = 64.0\%\end{tabular} \\
\multicolumn{1}{l}{} & \multicolumn{1}{l}{} & \multicolumn{1}{l}{} &  \\
\begin{tabular}[c]{@{}c@{}}Meng \textit{et al.}\\ \cite{meng2019machine} (2019)\end{tabular} & \begin{tabular}[c]{@{}c@{}}Health status \\ detection over time\end{tabular} & \begin{tabular}[c]{@{}c@{}}14 features acquired from activity tracker data to \\ feed different ML algorithms.\end{tabular} & AUC Score = 76\% \\
\multicolumn{1}{l}{} & \multicolumn{1}{l}{} & \multicolumn{1}{l}{} &  \\
\begin{tabular}[c]{@{}c@{}}Bent \textit{et al.}\\ \cite{bent2021engineering} (2021)\end{tabular} & \begin{tabular}[c]{@{}c@{}}Interstitial glucose \\ prediction\end{tabular} & \begin{tabular}[c]{@{}c@{}}69 physiological features acquired from smartwatches, diet, and \\ other biological data. Utilization of gradient-boosted techniques.\end{tabular} & Accuracy = 87\% \\
\multicolumn{1}{l}{} & \multicolumn{1}{l}{} & \multicolumn{1}{l}{} &  \\
\begin{tabular}[c]{@{}c@{}}Kim \textit{et al.}\\ \cite{kim2021building} (2021)\end{tabular} & \begin{tabular}[c]{@{}c@{}}Cardiovascular \\ diseases detection\end{tabular} & \begin{tabular}[c]{@{}c@{}}Features extracted from smartwatches and\\ user-related data to feed different ML techniques.\end{tabular} & \begin{tabular}[c]{@{}c@{}}Precision = 89.93\%\\ Recall = 85.59\%\\ F1 Score = 87.70\%\end{tabular} \\
\multicolumn{1}{l}{} & \multicolumn{1}{l}{} & \multicolumn{1}{l}{} &  \\
\begin{tabular}[c]{@{}c@{}}Bogue-Jimenez \textit{et al.}\\ \cite{bogue2022selection} (2022)\end{tabular} & \begin{tabular}[c]{@{}c@{}}Blood glucose \\ concentration prediction\end{tabular} & \begin{tabular}[c]{@{}c@{}}BGL detection using different ML techniques from \\ physiological features acquired from wristband-like devices.\end{tabular} & RMSE = 8.01\% \\
\multicolumn{1}{l}{} & \multicolumn{1}{l}{} & \multicolumn{1}{l}{} &  \\
\begin{tabular}[c]{@{}c@{}}Sadeghi \textit{et al.}\\ \cite{sadeghi2022posttraumatic} (2022)\end{tabular} & \begin{tabular}[c]{@{}c@{}}Hyperarousal \\ events detection\end{tabular} & \begin{tabular}[c]{@{}c@{}}Different ML algorithms applied to physiological \\ features acquired from wearable devices.\end{tabular} & \begin{tabular}[c]{@{}c@{}}Accuracy = 83\%\\ AUC Score = 70\%\end{tabular} \\
\multicolumn{1}{l}{} & \multicolumn{1}{l}{} & \multicolumn{1}{l}{} &  \\
\begin{tabular}[c]{@{}c@{}}Himi \textit{et al.}\\ \cite{himi2023medai} (2023)\end{tabular} & Diseases detection & \begin{tabular}[c]{@{}c@{}}Data acquired from anthropometric measurements, BGLs, \\ heart-related data, and sleep-related data to feed RF algorithm.\end{tabular} & Accuracy = 99.4\% \\ \hline
\end{tabular}%
}
\end{table}

Many studies have focused on analyzing PA through wearable devices, showing their effectiveness in promoting PA and reducing sedentary lifestyles \cite{ferguson2022effectiveness, yen2022smart, kim2019impact}. For example, a study by Cadmus \textit{et al.} \cite{cadmus2015use} involving FitBit trackers demonstrated high adherence levels to self-monitoring interventions. In glycemic health, wearable devices have been used for blood glucose levels (BGL) prediction. Approaches using invasive and non-invasive CGM devices have shown promising results in predicting hyper- or hypoglycemic events \cite{woldaregay2019data, deng2021deep}. For instance, Duckworth \textit{et al.} utilized CGM devices for hypo- and hyper-glycemia predictions using explainable ML algorithms \cite{duckworth2024explainable}. Bogue-Jimenez \textit{et al.} \cite{bogue2022selection} estimated BGL from wristband-like devices that collected physiological measurements, and data from electrodermal activity (EDA) and photoplethysmography (PPG) sensors to feed 8 different ML algorithms in the BGL detection. In \cite{bent2021engineering}, Bent \textit{et al.} utilized 69 features from distinct sources such as smartwatches, diet, and other physiological and biological parameters to predict interstitial glucose, demonstrating the feasibility of non-invasive methods.

Heart-related monitoring using wearable devices has also been extensively studied. However, this monitoring is typically measured in controlled scenarios (i.e., hospitals) and requires a follow-up intervention. Dunn \textit{et al.} \cite{dunn2021wearable} predicted clinical laboratory test results using data from wearable devices and feeding them into ML algorithms, showing more consistent and precise data than clinical measurements. Similarly, Kim \textit{et al.} \cite{kim2021building} developed an ML model for cardiovascular disease detection using smartwatch data. Concretely, these devices provided HR and oxygen saturation data, as well as stress markers obtained from breathing and body temperature. Meng \textit{et al.} \cite{meng2019machine} utilized activity tracker data to detect health status over time in ischemic heart disease patients by using different ML algorithms. Finally, 14 distinct features were selected from activity tracker data, including physical parameters such as steps, distance, active minutes, light minutes, or sedentary minutes.

Wearable devices also effectively monitor and influence behavioral changes in SA, which can directly impact health outcomes and improve physical and mental disorders. Automatic sleep stage classification in non-controlled interventions was challenging before the AI era. Nevertheless, the spread of DL techniques has led to improvements in sleep-related tasks \cite{mahmood2022wearable,cheung2019examining}. For instance, Zhang \textit{et al.} \cite{zhang2018sleep} used recurrent neural networks (RNNs) for sleep staging with wearable devices, while Beattie \textit{et al.} \cite{beattie2017estimation} considered automated classifiers for sleep stage classification using data from wrist-worn devices. In this last study, they captured sleep-related data from a 3D accelerometer and an optical pulse PPG to extract 180 distinct features.

Other studies have explored the use of wearables in mental health and global health issues. For example, Sadeghi \textit{et al.} \cite{sadeghi2022posttraumatic} developed a method to detect hyperarousal events in post-traumatic stress disorder individuals using wearable device data. Specifically, they acquired physiological data (i.e., HR and body acceleration) and self-reported hyperarousal events to finally feed them to ML algorithms such as Random Forest (RF), Support Vector Machine (SVM), and Logistic Regression (LR). During the COVID-19 pandemic, studies analyzed heart-related and PA data from fitness trackers, finding associations between abnormal resting HR and the illness \cite{lukas2020emerging, mishra2020pre}. Himi \textit{et al.} \cite{himi2023medai} finally presented a health mobile application that predicts multiple diseases using an RF algorithm. Data were collected from multiple sources, including anthropometric measurements, BGLs, and heart- and sleep-related data.

Previous studies have utilized wearable device data to predict various health outcomes, mainly relying on basic external features. However, none have specifically focused on predicting weight loss. Our study focuses on applying AI techniques to data from wearables, offering a novel approach in this area.

% \usepackage{multirow}
% \usepackage{graphicx}
\begin{table}[t]
\caption{Statistics of AI4FoodDB subjects, including anthropometric, biochemical, and dataset-specific metrics. We report the mean and standard deviation values of key features across the total population (93 subjects), subjects who lost $\geq$ 2\% of their initial weight (55 subjects), and subjects who did not (38 subjects). P-values were obtained via a Wilcoxon test for continuous variables and a Chi-square test for sex. Additionally, P-values were corrected via the Benjamin-Hochberg False Discovery Rate (FDR) correction.}
\label{tab:statistics}
\resizebox{\textwidth}{!}{%
\begin{tabular}{clccccc}
\hline
Dataset & Feature & \begin{tabular}[c]{@{}c@{}}Total \\ (n=93)\end{tabular} & \begin{tabular}[c]{@{}c@{}}Weight Loss $\geq$ 2\%\\ (n=55)\end{tabular} & \begin{tabular}[c]{@{}c@{}}Weight Loss $<$ 2\%\\ (n=38)\end{tabular} & P-value & \begin{tabular}[c]{@{}c@{}}Adjusted\\ P-value\end{tabular} \\ \hline
 & Age  & 50 $\pm$ 13 & 52 $\pm$ 12  & 45 $\pm$ 13     & $<$ 0.05   & 0.12            \\
&Sex (\% women)   & 70 \%  & 73 \%       & 65 \%      & 0.4     & 0.789              \\
Anthropometric &Body Mass Index (kg$/m^2$)   & 30.73 $\pm$ 3.36  & 30.90  $\pm$ 3.35 & 30.80 	 $\pm$  3.41 & 0.5     & 0.789              \\
and &Waist-Hip Ratio  & 0.88 $\pm$ 0.10  & 0.89 	$\pm$ 0.10 & 0.88 	 $\pm$ 0.10 & 0.6     & 0.789             \\
Biochemical &Systolic Blood Pressure (mmHg) & 124 $\pm$ 17 & 125 	 $\pm$ 19   & 123 		 $\pm$ 15   & 0.6     & 0.789             \\
Statistics&Diastolic Blood Pressure (mmHg) & 78 $\pm$ 10 & 78 	 $\pm$ 10    & 78 		 $\pm$ 9  & 0.4     & 0.789             \\
&High-Density Lipoprotein (mg/dL)  &  59 $\pm$ 14 & 58 	 $\pm$ 13    & 61  $\pm$ 15      & 0.2     & 0.615            \\
&Triglycerides (mg/dL) &  107 $\pm$ 46   & 111  $\pm$ 53    & 100 $\pm$ 35    & 0.5     & 0.789            \\  \hline

\multirow{4}{*}{DS4 (Biomarkers)} & Glucose in mg/dL & 100.05 $\pm$ 7.25 & 100.87 $\pm$ 8.06 & 98.85 $\pm$ 5.76 & 0.188 & 0.615 \\
 & HB1Ac & 5.11 $\pm$ 0.25 & 5.14 $\pm$ 0.28 & 5.07 $\pm$ 0.20 & 0.188 & 0.615 \\
 & Glucose variability (coefficient of variation) & 15.85 $\pm$ 3.71 & 16.64 $\pm$ 3.97 & 14.68 $\pm$ 2.96 & 0.011 & 0.137 \\
 & \% time in target values (70-180 mg/dL) & 98.01 $\pm$ 3.6 & 97.72 $\pm$ 4.1 & 98.44 $\pm$ 2.8 & 0.347 & 0.789 \\ \hline
\multirow{5}{*}{DS6 (Vital Signs)} & Heart rate in b.p.m. & 75.74 $\pm$ 6.42 & 75.81 $\pm$ 6.49 & 75.63 $\pm$ 6.42 & 0.896 & 0.946 \\
 & Resting heart rate & 62.84 $\pm$ 7.45 & 62.78 $\pm$ 7.34 & 62.93 $\pm$ 7.72 & 0.922 & 0.946 \\
 & Heart rate during physical activity & 101.23 $\pm$ 9.52 & 100.81 $\pm$ 9.22 & 101.84 $\pm$ 10.03 & 0.612 & 0.789 \\
 & Heart rate during non-REM sleep & 60.03 $\pm$ 7.94 & 59.56 $\pm$ 8.01 & 60.72 $\pm$ 7.89 & 0.494 & 0.789 \\
 & Heart rate during Electrocardiogram session & 69.93 $\pm$ 9.16 & 69.76 $\pm$ 9.28 & 70.22 $\pm$ 9.09 & 0.828 & 0.920 \\ \hline
\multirow{9}{*}{DS7 (Physical Activity)} & Calories & 2,983 $\pm$ 459 & 3,019 $\pm$ 389 & 2,932 $\pm$ 432 & 0.371 & 0.838 \\
 & Steps & 11,051 $\pm$ 3,760 & 11,356 $\pm$ 3,800 & 10,601 $\pm$ 4,145 & 0.342 & 0.789 \\
 & Nº physical activities performed & 14.89 $\pm$ 10 & 14.20 $\pm$ 10 & 15.92 $\pm$ 10 & 0.429 & 0.789 \\
 & Duration of physical activities in minutes & 37.73 $\pm$ 17.07 & 38.78 $\pm$ 19.39 & 36.19 $\pm$ 12.99 & 0.472 & 0.789 \\
 & Sedentary minutes & 12h 00min $\pm$ 1h 39min & 11h 57min $\pm$ 1h 35min & 12h 04min $\pm$ 1h 45min & 0.725 & 0.853 \\
 & \% days with $\geq$ 10 lightly active minutes & 99.75 $\pm$ 2.38 & 99.59 $\pm$ 3.08 & 100.00 $\pm$ 0.00 & 0.413 & 0.789 \\
 & \% days with $\geq$ 10 moderately active minutes & 68.27 $\pm$ 21.45 & 68.95 $\pm$ 23.25 & 67.28 $\pm$ 18.74 & 0.866 & 0.853 \\
 & \% days with $\geq$ 10 very active minutes & 65.52 $\pm$ 25.65 & 65.98 $\pm$ 27.31 & 64.86 $\pm$ 23.34 & 0.912 & 0.946 \\
 & MVPA minutes & 66.11 $\pm$ 42.27 & 68.18 $\pm$ 49.41 & 63.05 $\pm$ 29.08 & 0.566 & 0.789 \\ \hline
\multirow{9}{*}{DS8 (Sleep Activity)} & Oxygen saturation during sleep & 94.08 $\pm$ 1.25 & 94.00 $\pm$ 1.30 & 94.20 $\pm$ 1.18 & 0.456 & 0.789 \\
 & Sleep duration & 7h 01min $\pm$ 0h 52min & 6h 54min $\pm$ 0h 54min & 7h 11min $\pm$ 0h 46min & 0.109 & 0.483 \\
 & Awake duration & 0h 54min $\pm$ 0h 12min & 0h 55min $\pm$ 0h 12min & 0h 53min $\pm$ 0h 11min & 0.658 & 0.822 \\
 & Light sleep duration & 3h 49min $\pm$ 0h 37min & 3h 47min $\pm$ 0h 34min & 3h 52min $\pm$ 0h 40min & 0.565 & 0.789 \\
 & Deep sleep duration & 1h 01min $\pm$ 0h 14min & 0h 58min $\pm$ 0h 15min & 1h 04min $\pm$ 0h 12min & 0.059 & 0.295 \\
 & REM sleep duration & 1h 17min $\pm$ 0h 19min & 1h 14min $\pm$ 0h 19min & 1h 22min $\pm$ 0h 18min & 0.034 & 0.260 \\
 & Sleep score & 74.82 $\pm$ 4.27 & 74.10 $\pm$ 4.78 & 75.80 $\pm$ 3.15 & 0.046 & 0.260 \\
 & Weekdays sleep score & 74.68 $\pm$ 4.45 & 73.91 $\pm$ 4.92 & 75.89 $\pm$ 3.42 & 0.042 & 0.260 \\
 & Weekend days sleep score & 75.10 $\pm$ 4.83 & 74.86 $\pm$ 4.87 & 75.47 $\pm$ 4.81 & 0.552 & 0.789\\ \hline
\multirow{4}{*}{DS9 (Emotional State)} & Stress score & 77.29 $\pm$ 4.54 & 75.70 $\pm$ 4.64 & 79.84 $\pm$ 3.04 & $<$ 0.05 & 0.12 \\
 & Responsiveness points & 22.64 $\pm$ 2.51 & 21.87 $\pm$ 2.21 & 23.87 $\pm$ 2.54 & 0.782 & 0.894 \\
 & Exertion points & 23.59 $\pm$ 2.47 & 23.50 $\pm$ 2.79 & 23.73 $\pm$ 1.93 & 0.156 & 0.615 \\
 & Sleep points & 31.06 $\pm$ 4.05 & 30.33 $\pm$ 4.56 & 32.24 $\pm$ 2.84 & $<$ 0.05 & 0.137 \\ \hline
\end{tabular}%
}
\end{table}

\section{AI4FoodDB Database}\label{sec:datasets}

The AI4Food database (AI4FoodDB) was developed from a 1-month randomized controlled trial (RCT) involving 100 overweight and obese subjects who were monitored during a nutritional intervention. AI4FoodDB comprises 10 distinct datasets (DS), ranging from biological samples to continuous digital measurements, obtained from three different types of data acquisition (i.e., manual, clinical, and digital).

In the present study, we consider digital data from datasets DS4 (Biomarkers), DS6 (Vital Signs), DS7 (Physical Activity), DS8 (Sleep Activity), and DS9 (Emotional State). Subjects wore a Freestyle Libre 2 continuous glucose monitor (CGM) device to track their BGL (in mg/dL) every 15 minutes during the intervention. In addition, they wore a FitBit Sense smartwatch to track their physiological data and lifestyle habits such as PA, SA, and stress. Importantly, digital measurements were collected exclusively during a two-week period of the RCT. For detailed information about the AI4FoodDB database and the RCT, please refer to \cite{romero2023ai4fooddb}.

Table \ref{tab:statistics} provides descriptive statistics of AI4FoodDB’s subjects, showing the mean and standard deviation values of anthropometric and biochemical statistics, and some of the most important features from each dataset. We report the statistics for the total population (93 subjects), subjects who lost $\geq$ 2\% of their initial weight (55 subjects), and subjects who did not (38 subjects). This threshold was established by expert nutritionists based on observed weight loss patterns and the duration of the RCT, considering it a valid indicator of meaningful weight reduction and reflecting a clinically significant change within the study period. P-values are also presented to indicate the statistical significance of differences between both groups. 

Overall, the primary differences between both groups were observed in glucose variability and emotional state metrics (stress and sleep points). However, many of these differences did not retain statistical significance. These findings suggest no statistically significant differences between the groups despite some observable trends.

\section{Methods}\label{sec:methods}

This section describes the method proposed to predict weight loss using information extracted from wearable devices and the application of ML models. In particular, Sec.~\ref{ssec_features} describes the novel set of features proposed in the present study. Sec.~\ref{ssec_features_analysis} provides a statistical analysis of these features whereas Sec.~\ref{ssec_features_selection} details the feature selection algorithms studied. Finally, Sec.~\ref{ssec_features_selection} provides a description of the ML classifiers analyzed.

\subsection{Proposed Features}\label{ssec_features}

The data used for feature extraction were obtained from the AI4FoodDB during the digital intervention, i.e., when subjects wore wearable devices. Specifically, each individual's device tracked approximately 14 days of behavioral information, including continuous data (e.g., heart rate and blood glucose levels) and discrete data (e.g., hours of sleep and minutes of light active). For additional details about the data acquisition process, please refer to \cite{romero2023ai4fooddb}. Once the raw data are extracted from each subject's device, a preprocessing step is carried out. This includes a standardization process, where the raw data are filtered and formatted consistently, and the removal of unnecessary information. After preprocessing, feature extraction is performed on the processed data, proposing a novel set of 284 features from five datasets. We briefly describe the features extracted for each dataset below (for more details, we refer the reader to Tables \ref{tab:features} and \ref{tab:features2} in Appendix \ref{appendixA}):

\begin{itemize}
    \item \textit{Dataset 4 - Biomarkers:} The first 65 features are directly related to glucose levels and include five different subfeatures corresponding to five parts of the day: all day, morning (6h-12h), afternoon (12h-18h), evening (18h-24h), and night (0h-6h). These features are either directly extracted from blood sugar levels or derived from equations (e.g., HB1Ac average and glucose variability). Descriptive statistical features (e.g., average, standard deviation, variance, maximum, minimum, and range) are included from features 1 to 30. Features 31 to 65 represent the percentage of glucose levels in different ranges: very high values ($>$ 250 mg/dL),  high values (181-250 mg/dL), target values (70-180 mg/dL), low values (54-69 mg/dL), and very low values ($<$ 54 mg/dL).

    \item \textit{Dataset 6 - Vital Signs:} Features from 66 to 123 (58 features) describe HR and Electrocardiogram (ECG) sessions. Similar to Dataset 4, the first 30 features provide descriptive statistical information about HR, divided into five parts of the day. Eight features correspond to the average and standard deviation of HR-related measurements during various times of the day (e.g.,  resting, during PA, during non-REM sleep, and the root mean square of successive differences (RMSSD) during sleep). The remaining 20 features include information about HR from EDA sessions and the ECG session waveform slopes.

    \item \textit{Dataset 7 - Physical Activity:} Features from 124 to 167 (44 features) are related to the PA during the intervention. Key features include calories, steps, and minutes in different HR zones (fat burn, cardio, and peak) and PA levels (sedentary, lightly, moderately, and very active).     Additional features include the percentage of days with 10+ minutes of various activity levels, moderate to vigorous physical activity (MVPA) minutes, and similar features related to PA.

    \item \textit{Dataset 8 - Sleep Activity:} A total of 97 sleep-related features are extracted. General features include information about oxygen saturation, temperature, breathing rate across sleep stages, and detailed sleep scores from FitBit (composition, revitalization, and duration subscores). Specific features cover sleep efficiency, sleep start and end time, night awakenings, regular wake-up and bedtime, and the individuals' restlessness.
    
    \item \textit{Dataset 9 - Emotional State:} This dataset includes 20 features related to mental health. The first eight features provide information about the FitBit stress score, comprising sleep, responsiveness, and exertion points. The remaining features include descriptive statistics of skin conductance levels (SCL) from EDA sessions.
\end{itemize}

\subsection{Features Analysis}\label{ssec_features_analysis}

\begin{figure*}[t]
    \centering
    \includegraphics[width=0.98\linewidth]{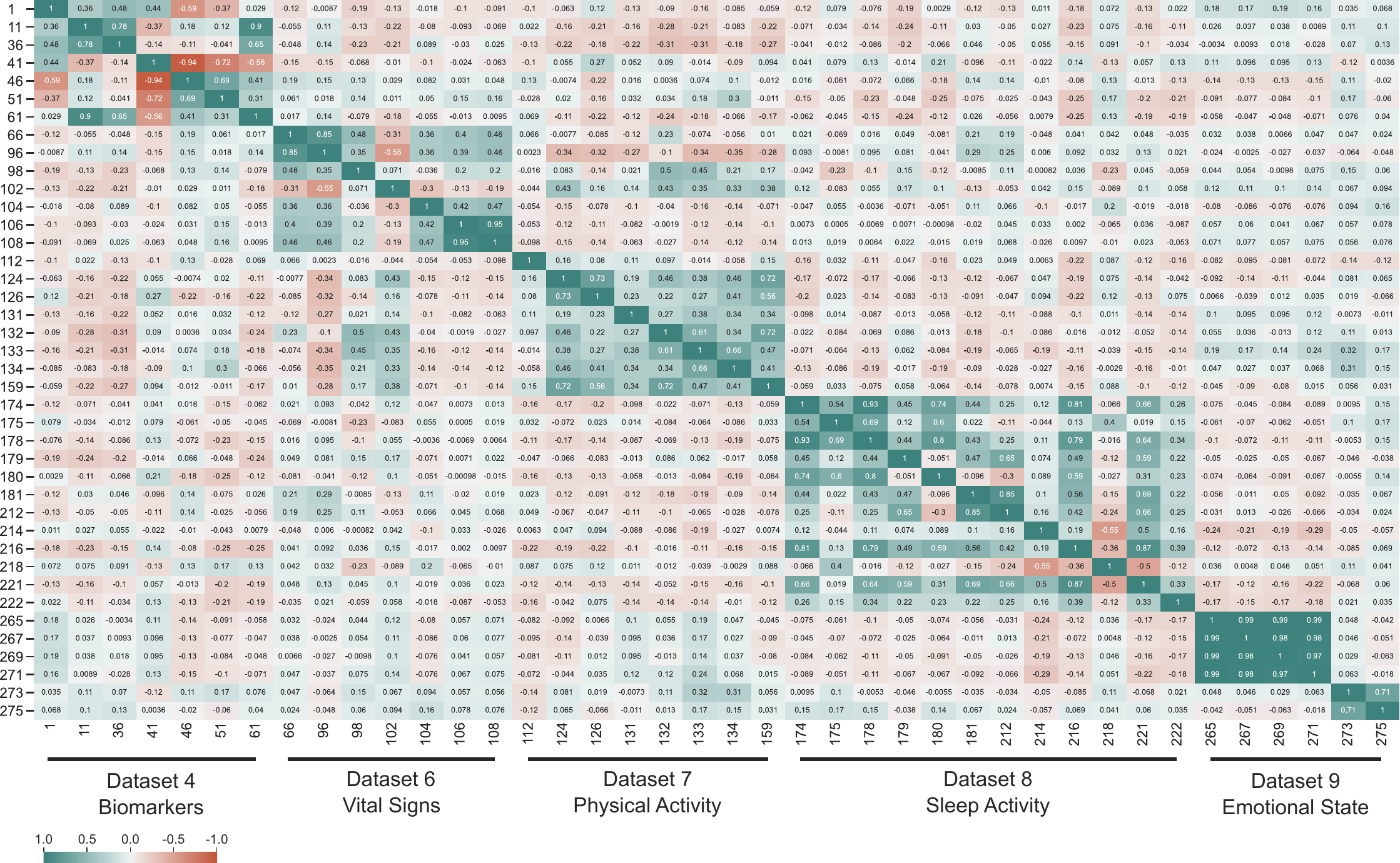}
    \caption{Correlation matrix generated by the Pearson correlation coefficient. Positive correlations are marked in \textbf{\textcolor{turquoise}{turquoise}} gradient, while negative correlations are marked in \textbf{\textcolor{red}{red}} gradient. The matrix reveals numerous strong correlations among features within the same dataset, indicating a linear relationship, but fewer correlations among features from different datasets.}
    \label{fig:corr_total}
\end{figure*}

Multimodal data from different sources and types are often noisy and not always comparable. Feature extraction, which transforms general into specific information, yields a better data representation. Additionally, normalized and similarly scaled data can reveal correlations among different datasets. In this study, we use the Pearson's correlation coefficient (PCC), a widely used correlation measure to assess the linear correlation between features, as shown in Eq. \ref{eq:pearson}:

\begin{equation}
    \rho_{f_1,f_2} = \frac{COV(f_1, f_2)}{\sigma_{f_1} \sigma_{f_2}}
    \label{eq:pearson}
\end{equation}

Given two features $f_{1}$ and $f_{2}$, the covariance between them is divided by the product of their standard deviations ($\sigma_{f_{1}}$ and $\sigma_{f_{2}}$). The resulting $\rho_{f_{1},f{_2}}$ ranges from $-1$ to $+1$, indicating the correlation degree. A negative PPC ($< 0$) indicates a negative linear correlation between $f_{1}$ and $f_{2}$, whereas a positive PPC ($> 0$) implies a positive linear correlation. A value of 0 means no correlation. Absolute values greater than 0.6 indicate a strong correlation whereas higher than 0.8 indicate an extremely strong correlation. 

In this study, we compute the PPC of all the proposed 284 features. Fig. \ref{fig:corr_total} shows the Pearson's correlation matrix for the most representative features (the feature numbers correspond to those in Tables \ref{tab:features} and \ref{tab:features2} in Appendix \ref{appendixA}). Positive and negative PPCs are marked in gradients of \textbf{\textcolor{turquoise}{turquoise}} and \textbf{\textcolor{red}{red}}, respectively. We observe numerous strong correlations among features within the same dataset, indicating linear relationships. For instance, positive correlations are seen between the variance of glucose and times in high values (DS4 - Features 11 and 36), HR and resting HR (DS6 - Features 66 and 96), calories and MVPA minutes (DS7 - Features 124 and 159), asleep minutes and duration score (DS8 - Features 174 and 216), and stress scores and related points (DS9 - Features 265, 267, 269, and 271). Conversely, a strong negative correlation is found between times in target values and the times in low and very low values (DS4 - Features 41, 46, and 51, respectively), showing an inverse proportionality.

However, there is no evidence of strong correlations between features from different datasets, which could be due to several factors. Relationships between these features might be weak and non-linear, while features from the same dataset present strong linear correlations. Different data acquisition devices could introduce variability and noise. Additionally, individual differences in physiological responses and lifestyle factors might result in inter-subject variability, complicating the detection of correlations.

\subsection{Feature Selection}\label{ssec_features_selection}

This section describes the feature selection techniques used to identify the most discriminative features for the prediction of weight loss \cite{tolosana2015preprocessing,ruiz2024childci}. Many of the initial 284 features may not contribute to distinguishing between subjects who lost weight and those who did not, leading to noise in the data. Therefore, feature selection is necessary to choose those features that minimize the intra-class variability and maximize the inter-class distance. 

\begin{itemize}
    \item \textit{Sequential Forward Floating Search (SFFS)} is one of the most used feature selection techniques, starting with an empty set of features and adding one feature at a time if the current feature improves the model's performance \cite{ferri1994comparative}. The estimator uses a cross-validation score to choose the best set of features. 
    \item \textit{Boruta Selection (BS)} is an algorithm designed to identify all relevant features by leveraging RF as its classifier. Additionally, this algorithm incorporates feature correlations to add the most discriminative ones    \cite{kursa2010feature}.
    \item \textit{Genetic Algorithm (GA)} is inspired by the natural process of evolution. Concretely, GA utilizes stochastic processes mimicking genetic variation, recombination, and selection to iteratively optimize solutions through evolutionary mechanisms \cite{babatunde2014genetic}.
\end{itemize}

\subsection{Classification Algorithms}\label{ssec_classifiers}

\begin{itemize}

    \item  \textit{Random Forest (RF)} is an ensemble method using multiple decision trees, called estimators, to predict outcomes by combining their decisions.
    \item \textit{Logistic Regression (LR)} is a statistical method analyzing datasets with one or more independent variables to determine an outcome measured. LR estimates the probability that a given input point belongs to a certain class.
    \item \textit{Gradient Boosting (GB)} is an ensemble learning technique used for regression and classification tasks. It builds the model using decision trees and combines them to create a strong predictive model, with each new tree correcting errors made by previous trees.

\end{itemize}

% \usepackage{graphicx}
\begin{table}[htpb]
\centering
\caption{Results achieved considering different state-of-the-art feature selection techniques and classifiers. The performance is measured using the area under the curve (AUC) metric (\%). The best model for each dataset and feature selection method is highlighted in \textbf{bold}. RF = Random Forest, LR = Logistic Regression, GB = Gradient Boosting, SFFS = Sequential Forward Floating Search.
}
\label{tab:results_1}
\resizebox{0.5\linewidth}{!}{%
\begin{tabular}{lcccc}
\hline
\multicolumn{1}{c}{} & \textbf{SFFS} & \textbf{\begin{tabular}[c]{@{}c@{}}Boruta\\ Selection\end{tabular}} & \textbf{\begin{tabular}[c]{@{}c@{}}Genetic\\ Algorithm\end{tabular}} & \textbf{\begin{tabular}[c]{@{}c@{}}All \\ Features\end{tabular}} \\ \hline
 & \multicolumn{1}{l}{} & \multicolumn{1}{l}{} & \multicolumn{1}{l}{} & \multicolumn{1}{l}{} \\ \hline
\multicolumn{5}{l}{\textbf{DS4: Biomarkers}} \\ \hline
RF & 64.45 & 72.37 & 60.82 & 59.46 \\
LR & \textbf{69.86} & \textbf{74.69} & \textbf{62.00} & \textbf{67.45} \\
GB & 66.11 & 72.93 & 54.88 & 55.42 \\\hline
 & \multicolumn{1}{l}{} & \multicolumn{1}{l}{} & \multicolumn{1}{l}{} & \multicolumn{1}{l}{} \\ \hline

\multicolumn{5}{l}{\textbf{DS6: Vital Signs}} \\ \hline
RF & 68.20 & 75.83 & 55.93 & 59.39 \\
LR & 52.39 & 58.33 & 52.46 & 51.84 \\
\textbf{GB} & \textbf{71.61} & \textbf{76.86} & \textbf{64.09} & \textbf{63.57} \\\hline
 & \multicolumn{1}{l}{} & \multicolumn{1}{l}{} & \multicolumn{1}{l}{} & \multicolumn{1}{l}{} \\ \hline
 
\multicolumn{5}{l}{\textbf{DS7: Physical Activity}} \\ \hline
RF & \textbf{67.88} & \textbf{69.31} & 62.54 & 63.78 \\
LR & 55.41 & 61.24 & 57.86 & 55.02 \\
GB & 67.06 & 63.52 & \textbf{75.65} & \textbf{65.49} \\ \hline
 & \multicolumn{1}{l}{} & \multicolumn{1}{l}{} & \multicolumn{1}{l}{} & \multicolumn{1}{l}{} \\ \hline

 \multicolumn{5}{l}{\textbf{DS8: Sleep Activity}} \\ \hline
RF & \textbf{63.88} & \textbf{71.72} & \textbf{61.46} & \textbf{62.22} \\
LR & 54.36 & 60.77 & 56.26 & 54.83 \\
GB & 62.34 & 66.97 & 54.33 & 52.61 \\ \hline
 & \multicolumn{1}{l}{} & \multicolumn{1}{l}{} & \multicolumn{1}{l}{} & \multicolumn{1}{l}{} \\ \hline

\multicolumn{5}{l}{\textbf{DS9: Emotional State}} \\ \hline
RF & 52.54 & 51.89 & 42.88 & 47.71 \\
LR & 48.72 & 52.86 & 50.13 & 48.18 \\
GB & \textbf{57.51} & \textbf{53.73} & \textbf{55.32} & \textbf{57.83} \\ \hline
 & \multicolumn{1}{l}{} & \multicolumn{1}{l}{} & \multicolumn{1}{l}{} & \multicolumn{1}{l}{} \\ \hline

\multicolumn{5}{l}{\textbf{Combined Datasets}} \\ \hline
RF & 79.22 & 79.15 & \textbf{72.12} & \textbf{69.12} \\
LR & 65.88 & 72.06 & 71.05 & 67.72 \\
GB & \textbf{84.44} & \textbf{82.96} & 70.80 & 56.33 \\ \hline
\end{tabular}%
}
\end{table}

\section{Experimental Framework}\label{sec:experiments}

\subsection{Protocol}

The current study aims to predict whether a person has lost more than 2\% of their initial weight using only digital data acquired from wearable devices. This value was selected following nutrition guidelines for the specific duration of our intervention (1 month). From the total 93 final subjects, 55 (59.14\%) lost more than the target percentage, whereas the remaining 38 subjects (40.86\%) lost less than 2\% of the weight or even gained weight. 

Due to the limited size of the dataset, we employed the widely used leave-one-out cross-validation (LOO-CV) method to ensure a robust evaluation of our models. In this approach, each iteration of the cross-validation process leaves out one subject per class as the test set, while the remaining subjects are used for training  \cite{osipov2024molecular}. This process is repeated until every subject has been used as a test case at least once, ensuring that the model is evaluated on all available data points.

Feature selection was applied separately within each iteration (cross-validation fold and scenario) to integrate it into our ML pipeline. Specifically, at each iteration, the training data were fed into various feature selection methods, which identified the most relevant features for that particular subset. This approach ensures that feature selection follows the same validation procedures as the rest of the pipeline, reducing the risk of overfitting in our limited dataset. 

To further enhance the reliability and robustness of our validation procedure, we performed five independent runs of the experiments, maintaining the optimal hyperparameters for all models but varying the random seed in each run. This strategy helps account for potential variability introduced by random initialization. The final performance metrics were computed as the mean across these five runs, providing a more stable and generalizable estimation of the model’s predictive capabilities.

The experimental protocol is divided into six different scenarios regarding the dataset utilized in each one. Therefore, there are five scenarios corresponding to DS4 (Biomarkers), DS6 (Vital Signs), DS7 (Physical Activity), DS8 (Sleep Activity), and DS9 (Emotional State). The remaining scenario is the combination of all datasets. This analysis allows to measure the contribution of each independent source of information, and the combination of all of them.

\begin{table}[ht]
\centering
\caption{Overview of the top 25 features selected for the optimal configuration (Combined Datasets) using the Gradient Boosting classifier and the Sequential Forward Floating Search feature selection technique. The features include 8 from glucose levels (Dataset 4), 1 from heart rate data (Dataset 6), 6 from physical activity parameters (Dataset 7), and 10 from sleep features (Dataset 8). The number of the feature is defined between parentheses.}
\label{tab:best_features}
\resizebox{0.5\linewidth}{!}{%
\begin{tabular}{ll}
\hline
\multicolumn{1}{c}{\textbf{Dataset}} & \multicolumn{1}{c}{\textbf{Feature (\#)}} \\ \hline
 & \cellcolor[HTML]{EFEFEF}std of glucose in the afternoon (8) \\
 & std of glucose in the evening (9) \\
 & \cellcolor[HTML]{EFEFEF}\% time in high values all day (36) \\
 & \begin{tabular}[c]{@{}l@{}}\% time in high values \\ in the morning (37)\end{tabular} \\
 & \cellcolor[HTML]{EFEFEF}HB1Ac avg all day (56) \\
 & HB1Ac avg in the afternoon (58) \\
 & \cellcolor[HTML]{EFEFEF}glucose variability in the morning (62) \\
\multirow{-8}{*}{DS4} & glucose variability in the afternoon (63) \\ \hline
DS6 & \cellcolor[HTML]{EFEFEF}avg RMSSD during sleep (102) \\ \hline
 & std of calories (125) \\
 & \cellcolor[HTML]{EFEFEF}std of steps (127) \\
 & std of distance (129) \\
 & \cellcolor[HTML]{EFEFEF}avg sedentary minutes last week (138) \\
 & avg minutes below default zone 1 (146) \\
\multirow{-6}{*}{DS7} & \cellcolor[HTML]{EFEFEF}avg MVPA minutes last week (167) \\ \hline
 & std of oxygen saturation during sleep (168) \\
 & \cellcolor[HTML]{EFEFEF}\begin{tabular}[c]{@{}l@{}}avg upper bound oxygen \\ saturation during sleep (173)\end{tabular} \\
 & avg asleep duration (174) \\
 & \cellcolor[HTML]{EFEFEF}std of std of REM sleep breathing rate (201) \\
 & avg revitalization score (214) \\
 & \cellcolor[HTML]{EFEFEF}std of revitalization score (215) \\
 & avg total overall sleep score (220) \\
 & \cellcolor[HTML]{EFEFEF}avg weekdays overall sleep score (221) \\
 & avg total sleep end time (232) \\
\multirow{-10}{*}{DS8} & \cellcolor[HTML]{EFEFEF}avg weekdays sleep end time (233) \\ \hline
\end{tabular}%
}
\end{table}

\subsection{Results}
Table \ref{tab:results_1} provides the results achieved considering different state-of-the-art feature selection techniques and classifiers. An additional table with detailed experimental results can be found in Table \ref{tab:results_2} in Appendix \ref{appendixA}, where additional ML algorithms are used to compare various classifiers and feature selectors, including Support Vector Machine (SVM), Multilayer Perceptron (MLP), and K-Nearest Neighbors (KNN). The metric used is the Area Under the Curve (AUC) for all models (\%). We highlight the best performance for each scenario and feature selection technique in \textbf{bold}. As can be seen in Table~\ref{tab:results_1}, the best results are achieved when we combine all sources of information (i.e., the combined datasets scenario) with results over 80\% AUC for the GB classifier and SFFS (84.44\% AUC) and Boruta Selection (82.96\% AUC) techniques. Even when no feature selection was applied (i.e., \textit{All Features} experiments), the ML models inherently prioritized certain features during training by assigning higher importance to those that contributed most to the prediction task. The most influential features for the optimal configuration are further analyzed in Sec. \ref{sec:best_conf}.

\begin{figure}[htpb]
    \centering
    \includegraphics[width=0.5\textwidth]{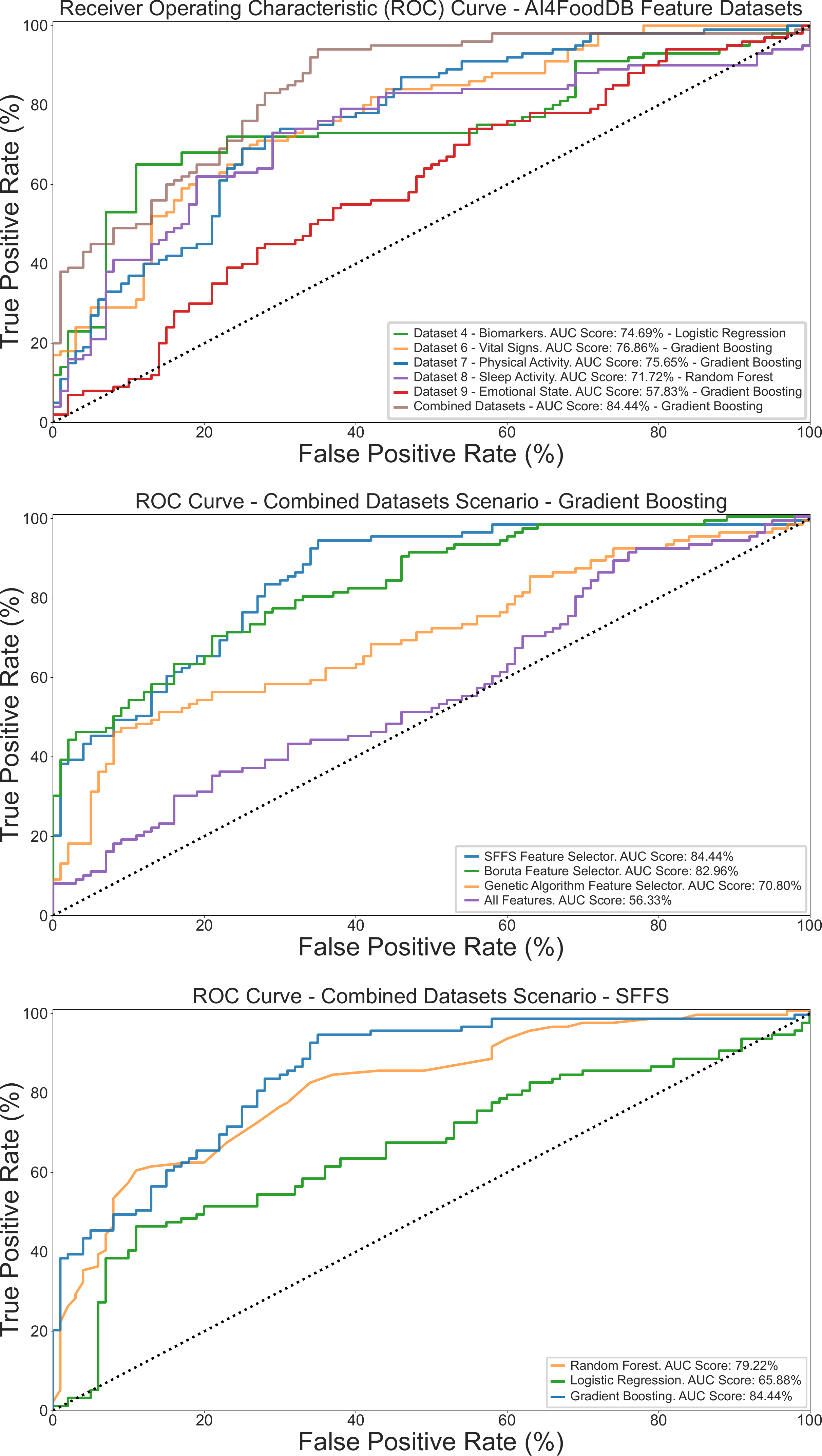}
    \caption{Receiver operating characteristic (ROC) curves for three different schemes. The first figure (top) shows the best configurations (feature selector and classifier) achieved in each scenario. The second figure (middle) includes the ROC curves for the Gradient Boosting (GB) models across different feature selectors for the combined datasets scenario. The third figure (bottom) shows the models for the combined datasets scenario using the Sequential Forward Floating Search (SFFS) feature selector. }
    \label{fig:ROC}
\end{figure}

To visually represent the performance of each scenario, Fig. \ref{fig:ROC} shows the receiver operating characteristic (ROC) curves with the bests configurations involving various feature selectors and classifiers. The curves plot the true positive rate (TPR, Sensitivity) against the false positive rate (FPR, 100-Specificity) for multiple classification thresholds. In the first figure (top), the ROC curve in the combined datasets scenario demonstrates again the best performance in general (84.44\% AUC), with a TPR that remains linear up to reaching 92\% for a 36\% FPR. Beyond this point, the curve increases slowly until reaching a TPR of around 98\% for an FPR of 60\%. Conversely, the ROC curve for DS9 (i.e., Emotional State) shows the lowest performance, characterized by instability from the beginning to the end. The second figure (middle) focuses on the GB models, comparing their performance across different feature selectors in the combined datasets scenario. The third figure (bottom) illustrates the models for the combined datasets scenario using the SFFS feature selector, where the LR achieves an AUC score of 65.88\%, and the RF model achieves an AUC score of 79.22\%. 

\begin{figure}
    \centering
    \includegraphics[width=0.98\linewidth]{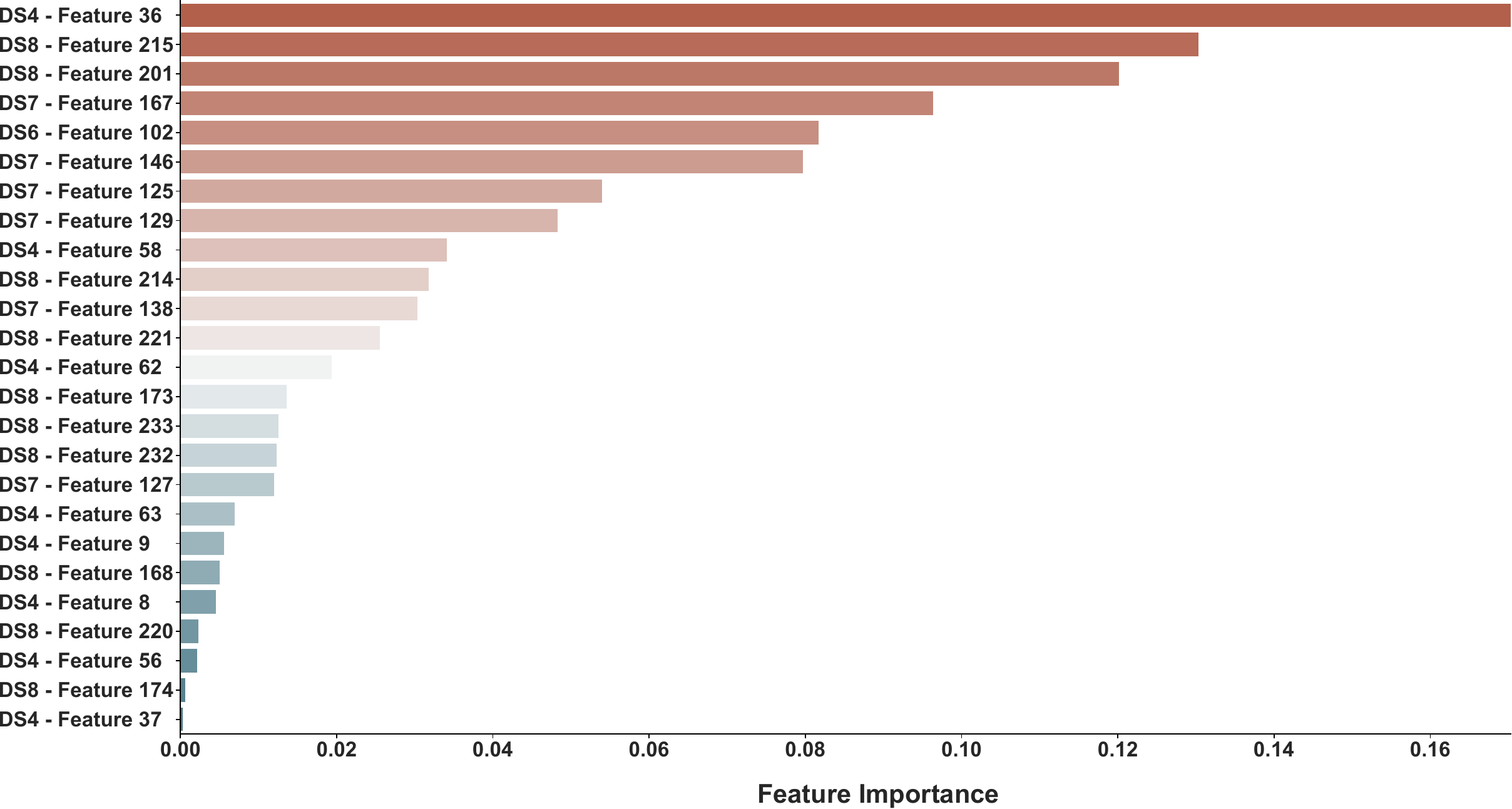}
    \caption{Feature importance of the 25 selected features in the Gradient Boosting (GB) classifier using the Sequential Forward Floating Search (SFFS) feature selector. The top three most influential features contribute to 42\% of the model's predictive power.}
    \label{fig:feature_importance}
\end{figure}

\subsection{Best Configuration Analysis}\label{sec:best_conf}
In the best configuration achieved (i.e., combined datasets scenario, using GB classifier and SFFS feature selector), from the original 284 features extracted, a total of 25 features were finally selected. These features, derived from multiple datasets, play a crucial role in differentiating between individuals who achieved a weight loss of at least 2\% and those who did not. Specifically, 8 features are derived from glucose levels (DS4), 1 from HR-related data (DS6), 6 from PA parameters (DS7), and the 10 remaining from sleep features (DS8). Table \ref{tab:best_features} describes the final 25 features and their respective datasets. Features from all datasets have been selected for this configuration, apart from the Emotional State dataset (DS9), highlighting the importance of combining multiple sources of information for the final decision.

To further analyze the impact of these 25 features, we examined their importance within the GB classifier. Fig. \ref{fig:feature_importance} presents the feature importance scores, grouped by dataset and feature type. The GB classifier identified three key features with an importance of $\geq 0.10$: DS4 - Feature 36 (\textit{\% time in high glucose values throughout the day}), DS8 - Feature 215 (\textit{standard deviation of revitalization score}), and DS8 - Feature 201 (\textit{double standard deviation of REM sleep breathing rate}). Individuals who achieved a weight loss of at least 2\%  exhibited higher levels of \textit{\% time in high glucose values throughout the day} and lower levels of \textit{standard deviation of revitalization score} and \textit{double standard deviation of REM sleep breathing rate}. Together, these three features contribute to 42\% of the final prediction, highlighting their substantial role in weight loss classification.

\begin{figure}
    \centering
    \includegraphics[width=0.98\linewidth]{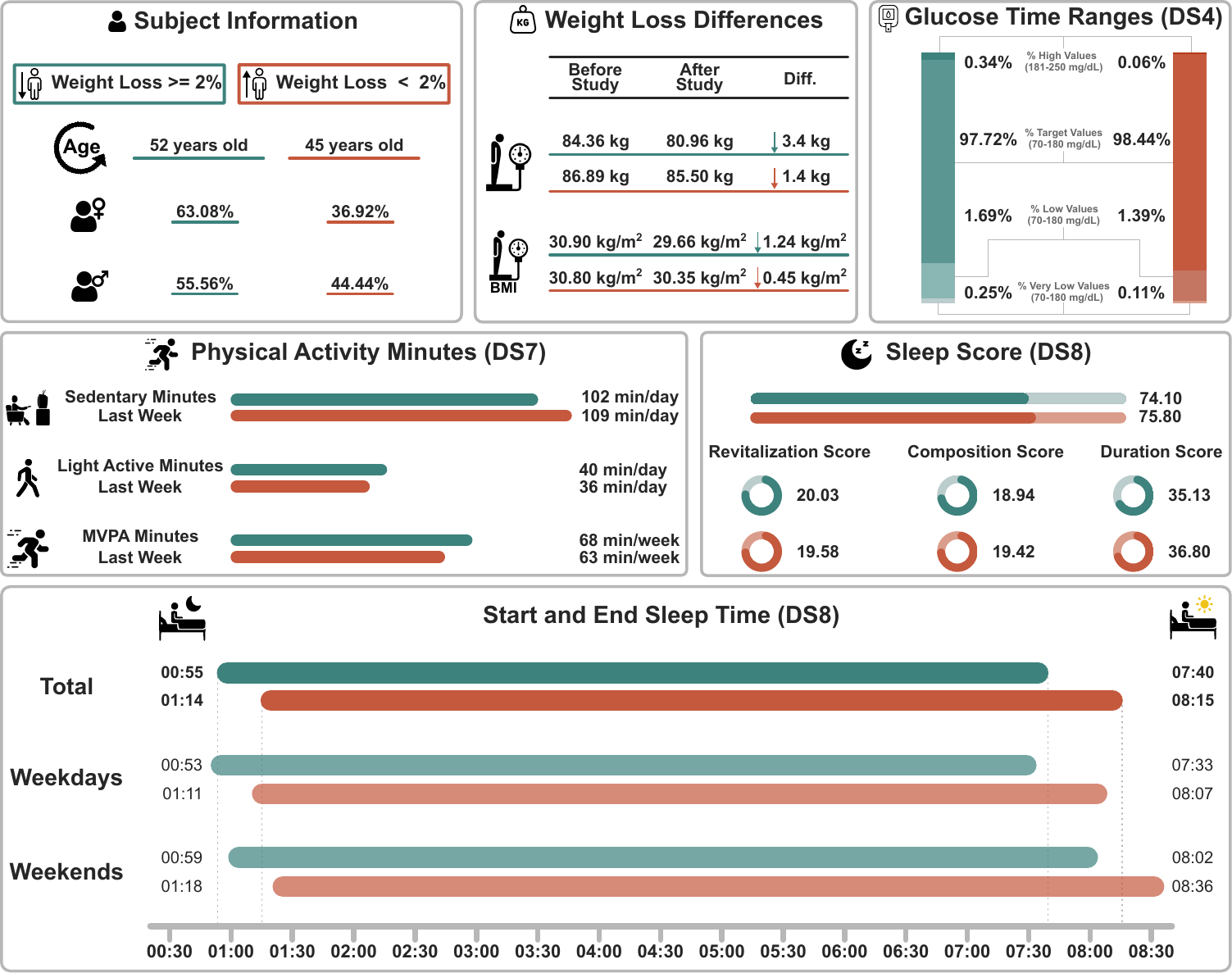}
    \caption{Comparative analysis of health metrics between participants who achieved significant weight loss ($>=2\%$) and those who did not ($<2\%$). Subject information and weight loss differences are also provided. Additionally, we include some specific feature information related to the best configuration achieved for DS4 (Biomarkers), DS7 (Physical Activity), and DS8 (Sleep Activity), described in detail in Table~\ref{tab:best_features}. }
    \label{fig:features}
\end{figure}

Given the short duration of the intervention, we think the impact of these features likely reflects short-term metabolic and physiological responses. Higher glucose levels may indicate individuals more responsive to dietary changes (DS4 - Feature 36), while lower variability in revitalization and REM sleep breathing rate suggests more stable physiological regulation (DS8 - Features 201 and 215), which could support adherence to weight loss strategies.

For completeness, we show in Fig. \ref{fig:features} a comparative analysis of various health metrics between both groups.  Although the statistical differences between the two groups for each feature are minimal, they still contribute significantly to the model's decision-making process. Concretely, some of the 25 selected features reflect differences between both groups. For instance, the DS4 - Feature 36 \textit{(\% time in high values all day)}, which was one of the most important features for the GB classifier, is almost 6 times higher in people who lost weight (0.34 vs 0.06\%). Features like DS7 - Feature 160 (\textit{avg sedentary minutes last week}) and DS7 - Feature 167 \textit{(avg MVPA minutes last week)} also show slight differences. People who lost more weight engaged in more MVPA minutes in the last seven days (68 vs 63 minutes per week) and had fewer sedentary minutes (102 vs 109 minutes per day). PA, primarily aerobic and resistance training, is part of the first-line treatments in weight loss interventions and has proven benefits beyond weight loss, contributing to a decreased risk of NCDs \cite{warburton_health_2006}. In general, both groups had similar SA patterns in terms of scores. However, those who did not lose weight tended to go to bed and wake up later, by about 20 and 35 minutes, respectively (DS8 - Feature 232, \textit{avg total sleep end time}). Being awake when circadian rhythms promote sleep, a phenomenon called circadian misalignment, has adverse effects on metabolic health and can contribute to obesity, even with small misalignments such as those caused by staying up late \cite{chaput_role_2023}.  Finally, demographic differences also play a significant role in this intervention. Individuals who lost weight were older (52 vs 45 years old), and a higher percentage of women lost weight compared to men (63.08\% vs 55.56\%). Furthermore, those who lost weight had a higher initial BMI compared to those who did not (30.90 vs 30.80 kg/$m^2$).

\subsection{Case Study}
\begin{figure}
    \centering
    \includegraphics[width=0.98\linewidth]{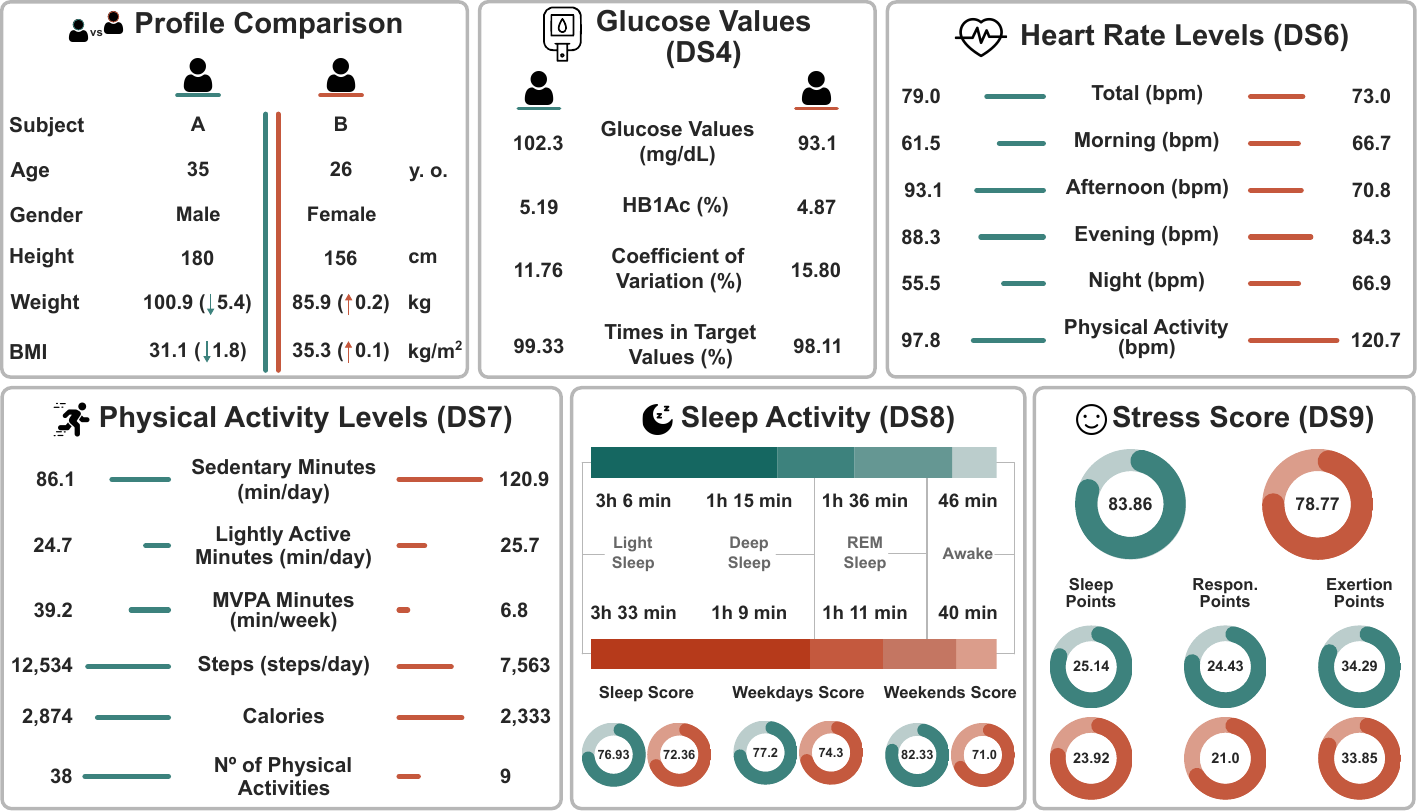}
    \caption{Comparative case study of two subjects: Subject A (35-year-old male, lost 5.4 kg) and Subject B (26-year-old female, gained 0.2 kg). General information (age, gender, height, etc.) and specific features are compared from the following datasets (DS): DS4 (Biomarkers), DS6 (Vital Signs), DS7 (Physical Activity), DS8 (Sleep Activity), and DS9 (Emotional State). Respon. Points = Responsiveness Points.}
    \label{fig:case_study}
\end{figure}

We finally analyze a case study of 2 different subjects in Fig.  \ref{fig:case_study}. Subject A was a 35-year-old male with an initial weight of 106.3 kg and a final weight of 100.9 kg. In contrast, subject B, who was a 26-year-old female, gained 0.2 kg (from 85.7 to 85.9 kg). In general, subject A had a higher average glucose value than subject B, but exhibited a more stable coefficient of variation (11.76\% vs 15.80 \%). Glucose variability can be used to assess glucose homeostasis, being in fact increased in prediabetic subjects \cite{bp_metrics_2017}. Regarding heart-related data, subject A shows a higher average HR level, but these values are much lower during the morning and night. Additionally, the pulse rate during PA (DS6 - Feature 98) was significantly lower (97.8 vs 120.7 bpm.). In terms of PA levels, subject A led a more active lifestyle: fewer sedentary minutes (86.1 vs 120.9 minutes per day , DS7 - Feature 138), more MVPA minutes (39.2 vs 6.8 minutes per week, DS7 - Feature 159), more steps (12,534 vs 7,563 steps, DS7 - Feature 126), and more physical activities performed (DS7 - Feature 130). In addition, although both subjects had similar SA levels, subject A had an average better performance in deep and REM sleep (approximately 6 and 25 minutes more in these sleep stages, DS8 - Features 179 and 181, respectively). Sleep scores were higher for the subject who lost more weight (76.93 vs 72.36 points, DS8 - Feature 220): subject A slept worse during the weekdays than on weekends (77.2 vs 82.33 points, DS8 - Features 221 and 222), whereas subject B  showed the opposite pattern (74.3 vs 71 points). Lastly, the stress score for subject A was also higher than for subject B (83.86 vs 78.77 points, DS9 - Feature 265), indicating a less stressed lifestyle in the subject who lost weight.

These findings underscore the potential of integrating wearable device data and AI to predict weight loss outcomes in overweight and obese individuals. The collective contribution of features related to glucose variability, PA, and SA forms a comprehensive profile that effectively distinguishes between those likely to achieve significant weight loss and those who are not. While individual feature differences may be minor,  their collective contribution enables the model to effectively predict weight loss outcomes.

\section{Discussion}
Existing literature has explored the integration of clinical data with information derived from wearable devices to predict various health outcomes. These studies often rely on simple external features extracted from these devices, which are sufficient to achieve their specific goals. However, to the best of our knowledge, no prior research has focused on predicting weight loss using data extracted solely from wearable devices in combination with AI methods. This gap is particularly relevant given the exponential surge in the adoption of smartwatches and other health-related devices in recent years. Addressing this limitation, our study is the first to integrate multiple physiological and behavioral features from wearable devices to predict weight loss using AI.

By predicting weight loss based on wearable-derived features, our ML model could facilitate the development of personalized health recommendations, potentially reducing the reliance on invasive metabolic assessments. While the database analysis reveals no robust, statistically significant differences between the groups, we could still observe and detect certain patterns distinguishing them. The non-linear nature of these physiological patterns explains why traditional linear methods, such as principal component analysis (PCA), proved inadequate for this task. These findings highlight the complexity of the problem, suggesting that multiple factors beyond wearable-derived data contribute to individual outcomes.

Numerous factors influence an individual's ability to manage weight, with digital data playing a crucial role in identifying societal patterns related to different lifestyles. However, digital data alone cannot fully distinguish individuals who achieve weight reduction solely through lifestyle changes. A more comprehensive understanding of the process could be achieved by integrating biological, environmental, and behavioral factors. This integration could enhance classifier performance and provide a more personalized and holistic view of the problem \cite{romero2024leveraging}. In this context, our study serves as a preliminary demonstration of the predictive power of wearable-derived data within broader multimodal research. Future work will integrate biological data from AI4FoodDB, such as biomarkers, metagenomics, and metabolomics, to enhance prediction accuracy and gain deeper insights into weight loss mechanisms, including the role of nutrition and metabolic responses. While the current study establishes a baseline utilizing accessible, low-cost, non-invasive wearable-derived features, emerging genomic biomarkers, like mitochondrial-derived peptides, may offer valuable contributions in future studies by providing the broader biological context necessary for state-of-the-art precision medicine  \cite{miller2022mitochondria}.

This study primarily concentrated on global features, while local features, such as time signals, remain largely unexplored. Investigating these local features in future research may prove crucial, as demonstrated in other domains like atrial fibrillation prediction \cite{MELZI2023100811, MARSILI2020103540}. A limitation in the current literature is the scarcity of real-world datasets similar to ours, which limits the generalizability of the findings. One potential solution is the use of synthetic data \cite{romero2023ai4food}. However, to ensure robustness and reduce bias, future studies should validate these findings on larger, more diverse populations. Additionally, exploring correlations between variables from different domains—such as glucose levels before and after meals, sleep patterns, HR, and stress levels before and after PA—could provide a deeper understanding of how these factors interact. Improving the interpretability of the model would assist healthcare professionals in understanding the decision-making process, leading to more effective personalized interventions.

\section{Conclusion}\label{sec:conclusion}
NCDs present a substantial challenge to contemporary society, impacting global health and economies. The exponential increase in wearable device usage and the integration of digital data are guiding us toward a future where continuous tracking of biomarkers can detect NCDs and other health-related conditions. Furthermore, recent advancements in AI enable personalized healthcare by detecting subtle changes from vast datasets for each individual. 

In this study, we have made an initial step towards detecting weight loss solely through data captured by wearable devices. Our results suggest that it is possible to differentiate between certain population groups with high performance, achieving over 80\% AUC in several scenarios using different ML models. The best results obtained indicate that using all datasets (combined datasets scenario) from the digital database is essential to better performance. Therefore, each subject's lifestyle markers must be considered to accurately identify those who lose weight, as these markers play a crucial role in the differentiation process.

\section{Acknowledgments}

This project has been supported by INSPIRACM (P2022/BMD7224), Cátedra ENIA UAM-VERIDAS en IA Responsable (NextGenerationEU PRTR TSI100927-2023-2), M2RAI (PID2024-160053OB-I00, MICIU/FEDER), TRUST-ID (PID2025-173396OB-I00 MICIU/AEI and the EU) and PowerAI+ (SI4/PJI/2024-00062 Comunidad de Madrid and UAM).

\bibliographystyle{elsarticle-num}
\bibliography{main}

\appendix
\section{}\label{appendixA}
The appendix includes three tables. Tables \ref{tab:features} and \ref{tab:features2} provide a comprehensive description of the digital data features proposed in the study. Table \ref{tab:results_2} presents the results of evaluating the AI4FoodDB database using seven state-of-the-art classifiers and feature selection methods, highlighting in \textbf{bold} the performance metrics for each approach.

\newpage

% \usepackage{graphicx}
\begin{table}[]
\renewcommand{\thetable}{\Roman{table}A}
\caption{Description of the digital features proposed in this study. Some features are divided into 5 different subfeatures, corresponding to different parts of the day: all day, morning (6h-12h), afternoon (12h-18h), evening (18h-24h), and night (0h-6h). Others are divided into subfeatures described in each of them. DS = Dataset.}
\label{tab:features}
\resizebox{\textwidth}{!}{%
\begin{tabular}{ccc|ccc}
\hline
\# & DS & Features & \# & DS & Features \\ \hline
1-5 & 4 & average (avg) glucose & 115 & 6 & \textit{max} of w.s. from ECG sessions \\
6-10 & 4 & standard deviation (\textit{std}) of glucose & 116 & 6 & \textit{min} of w.s. from ECG sessions \\
11-15 & 4 & glucose variance & 117 & 6 & \textit{min}-\textit{max} difference of w.s. from ECG sessions \\
16-20 & 4 & \textit{max}imum (\textit{max}) of glucose & 118 & 6 & \textit{avg} \textit{std} of w.s. from ECG sessions \\
21-25 & 4 & \textit{min}imum (\textit{min}) of glucose & 119 & 6 & \textit{std} of \textit{std} of w.s. from ECG sessions \\
26-30 & 4 & \textit{min}-\textit{max} difference of glucose & 120 & 6 & \textit{std} of w.s. from ECG sessions variance \\
31-35 & 4 & \% time in very high values ($\geq$ 250 mg/dL) & 121 & 6 & \textit{max} of \textit{std} of w.s. from ECG sessions \\
36-40 & 4 & \% time in high values (181-250 mg/dL) & 122 & 6 & \textit{min} of \textit{std} of w.s. from ECG sessions \\
41-45 & 4 & \% time in target values (70-180 mg/dL) & 123 & 6 & \begin{tabular}[c]{@{}c@{}}\textit{min}-\textit{max} difference of \textit{std} \\ of w.s. from ECG sessions\end{tabular} \\
46-50 & 4 & \% time in low values (54-69 mg/dL) & 124 & 7 & \textit{avg} calories \\
51-55 & 4 & \% time in very low values ($\le$ 54 mg/dL) & 125 & 7 & \textit{std} of calories \\
56-60 & 4 & HB1Ac \textit{avg} & 126 & 7 & \textit{avg} steps \\
61-65 & 4 & glucose variability (coefficient of variation) & 127 & 7 & \textit{std} of steps \\ 
66-70 & 6 & \textit{avg} heart rate (HR) & 128 & 7 & \textit{avg} distance \\
71-75 & 6 & \textit{std} of HR & 129 & 7 & \textit{std} of distance \\
76-80 & 6 & HR variance & 130 & 7 & \begin{tabular}[c]{@{}c@{}}number of physical \\ activities performed\end{tabular} \\
81-85 & 6 & \textit{max} of HR & 131 & 7 & \begin{tabular}[c]{@{}c@{}}\textit{avg} duration of physical \\ activities (\textit{min}utes)\end{tabular} \\
86-90 & 6 & \textit{min} of HR & 132-134 & 7 & \textit{avg} \{fat burn, cardio, peak\} \textit{min}utes \\
91-95 & 6 & \textit{min}-\textit{max} difference of HR & 135-137 & 7 & \textit{std} of \{fat burn, cardio, peak\} \textit{min}utes \\
96 & 6 & \textit{avg} resting HR & 138 & 7 & \textit{avg} sedentary \textit{min}utes \\
97 & 6 & \textit{std} of resting HR & 139 & 7 & \textit{std} of sedentary \textit{min}utes \\
98 & 6 & \textit{avg} HR during physical activity & 140-142 & 7 & \begin{tabular}[c]{@{}c@{}}\textit{avg} \{lightly, moderately, very\} \\ active \textit{min}utes\end{tabular} \\
99 & 6 & \textit{std} of HR during physical activity & 143-145 & 7 & \begin{tabular}[c]{@{}c@{}}\textit{std} of \{lightly, moderately, very\} \\ active \textit{min}utes\end{tabular} \\
100 & 6 & \textit{avg} HR during non-REM sleep & 146 & 7 & \textit{avg} \textit{min}utes below default zone 1 \\
101 & 6 & \textit{std} of HR during non-REM sleep & 147 & 7 & \textit{std} of \textit{min}utes below default zone 1 \\
102 & 6 & \begin{tabular}[c]{@{}c@{}}\textit{avg} root mean square of successive \\ differences (RMSSD) during sleep\end{tabular} & 148-150 & 7 & \textit{avg} \textit{min}utes in default zone \{1, 2, 3\} \\
103 & 6 & \textit{std} of RMSSD during sleep & 151-153 & 7 & \textit{std} of \textit{min}utes in default zone \{1, 2, 3\} \\
104 & 6 & \begin{tabular}[c]{@{}c@{}}\textit{avg} HR during Electrodermal Activity \\ (EDA) sessions\end{tabular} & 154 & 7 & \textit{avg} demographic VO$_2$ \textit{max} \\
105 & 6 & \textit{std} of HR during EDA sessions & 155 & 7 & \textit{std} of demographic VO$_2$ \textit{max} \\
106 & 6 & \textit{avg} HR at the beginning of EDA sessions & 156-158 & 7 & \begin{tabular}[c]{@{}c@{}}\% of days with $\geq$ 10 \\ \{lightly, moderately, very\} active \textit{min}/day\end{tabular} \\
107 & 6 & \textit{std} of HR at the beginning of EDA sessions & 159 & 7 & \begin{tabular}[c]{@{}c@{}}\textit{avg} moderate to vigorous \\ physical activity (MVPA) \textit{min}utes\end{tabular} \\
108 & 6 & \textit{avg} HR at the end of EDA sessions & 160 & 7 & \textit{avg} sedentary \textit{min}utes last week \\
109 & 6 & \textit{std} of HR at the end of EDA sessions & 161-163 & 7 & \begin{tabular}[c]{@{}c@{}}\textit{avg} \{lightly, moderately, very\}\\ active \textit{min}utes last week\end{tabular} \\
110 & 6 & \begin{tabular}[c]{@{}c@{}}\textit{avg} heart rate variability (HRV) \\ baseline during EDA sessions\end{tabular} & 164-166 & 7 & \begin{tabular}[c]{@{}c@{}}\% of days with $\geq$ 10 \\ \{lightly, moderately, very\} \\ active \textit{min}/day last week\end{tabular} \\
111 & 6 & \textit{std} of HRV baseline during EDA sessions & 167 & 7 & \textit{avg} MVPA \textit{min}utes last week \\
112 & 6 & \textit{avg} waveform slope (w.s.) from ECG sessions & 168 & 8 & \textit{avg} oxygen saturation during sleep \\
113 & 6 & \textit{std} of w.s. from ECG sessions & 169 & 8 & \textit{std} of oxygen saturation during sleep \\
114 & 6 & variance of w.s. from ECG sessions & 170 & 8 & \begin{tabular}[c]{@{}c@{}}\textit{avg} lower bound oxygen \\ saturation during sleep\end{tabular} \\ \hline
\end{tabular}%
}
\end{table}

% \usepackage{graphicx}
% \usepackage[normalem]{ulem}
\begin{table}[]
\addtocounter{table}{-1}
\renewcommand{\thetable}{\Roman{table}B}
        \caption{Continuation of Table \ref{tab:features}.}

\label{tab:features2}
\resizebox{\textwidth}{!}{%
\begin{tabular}{ccc|ccc}
\hline
\# & DS & Features & \multicolumn{1}{c}{\#} & DS & Features \\ \hline
171 & 8 & \begin{tabular}[c]{@{}c@{}}\textit{std} of lower bound oxygen \\ saturation during sleep\end{tabular} & \multicolumn{1}{c}{238-240} & 8 & \begin{tabular}[c]{@{}c@{}}\textit{avg} \{total, weekdays, weekend days\}\\ early waking up deviation time\end{tabular} \\
172 & 8 & \begin{tabular}[c]{@{}c@{}}\textit{avg} upper bound oxygen \\ saturation during sleep\end{tabular} & \multicolumn{1}{c}{241-243} & 8 & \begin{tabular}[c]{@{}c@{}}\textit{avg} \{total, weekdays, weekend days\}\\ late waking up deviation time\end{tabular} \\
173 & 8 & \begin{tabular}[c]{@{}c@{}}\textit{std} of upper bound oxygen \\ saturation during sleep\end{tabular} & \multicolumn{1}{c}{244-246} & 8 & \begin{tabular}[c]{@{}c@{}}\% of \{days, weekdays, weekend days\}\\ of regular wake-up\end{tabular} \\
174-175 & 8 & \textit{avg} \{asleep, awake\} duration (\textit{min}utes) & \multicolumn{1}{c}{247-249} & 8 & \begin{tabular}[c]{@{}c@{}}\% of \{days, weekdays, weekend days\}\\ of regular bedtime\end{tabular} \\
176-177 & 8 & \textit{std} \{asleep, awake\} duration (\textit{min}utes) & \multicolumn{1}{c}{250-252} & 8 & \begin{tabular}[c]{@{}c@{}}\% of \{days, weekdays, weekend days\}\\ restful sleep with over 25\% REM\end{tabular} \\
178-181 & 8 & \begin{tabular}[c]{@{}c@{}}\textit{avg} \{full, deep, light, REM\} \\ duration night sleep (\textit{min}utes)\end{tabular} & \multicolumn{1}{c}{253-255} & 8 & \begin{tabular}[c]{@{}c@{}}\% of \{days, weekdays, weekend days\}\\ early waking time\end{tabular} \\
182-185 & 8 & \begin{tabular}[c]{@{}c@{}}\textit{std} \{full, deep, light, REM\} \\ duration night sleep (\textit{min}utes)\end{tabular} & \multicolumn{1}{c}{256-258} & 8 & \begin{tabular}[c]{@{}c@{}}\% of \{days, weekdays, weekend days\}\\ late waking time\end{tabular} \\
186-189 & 8 & \begin{tabular}[c]{@{}c@{}}\textit{avg} \{full, deep, light, REM\} \\ sleep breathing rate\end{tabular} & \multicolumn{1}{c}{259-261} & 8 & \begin{tabular}[c]{@{}c@{}}\% of \{days, weekdays, weekend days\}\\ better restlessness variations\end{tabular} \\
190-193 & 8 & \begin{tabular}[c]{@{}c@{}}\textit{std} of \{full, deep, light, REM\}\\ sleep breathing rate\end{tabular} & \multicolumn{1}{c}{262-264} & 8 & \begin{tabular}[c]{@{}c@{}}\% of \{days, weekdays, weekend days\} \\ worse restlessness variations\end{tabular} \\
194-197 & 8 & \begin{tabular}[c]{@{}c@{}}\textit{avg} \textit{std} \{full, deep, light, REM\}\\ sleep breathing rate\end{tabular} & 265 & 9 & \textit{avg} stress score \\
198-201 & 8 & \begin{tabular}[c]{@{}c@{}}\textit{std} of \textit{std} of \{full, deep, light, REM\}\\ sleep breathing rate\end{tabular} & 266 & 9 & \textit{std} of stress score \\
202-205 & 8 & \begin{tabular}[c]{@{}c@{}}\textit{avg} \{full, deep, light, REM\}\\ sleep breathing rate signal to noise\end{tabular} & 267 & 9 & \textit{avg} sleep points \\
206-209 & 8 & \begin{tabular}[c]{@{}c@{}}\textit{std} of \{full, deep, light, REM\}\\ sleep breathing rate signal to noise\end{tabular} & 268 & 9 & \textit{std} of sleep points \\
210 & 8 & \textit{avg} nightly temperature & 269 & 9 & \textit{avg} responsiveness points \\
211 & 8 & \textit{std} of nightly temperature & 270 & 9 & \textit{std} of responsiveness points \\
212 & 8 & \textit{avg} composition score & 271 & 9 & \textit{avg} exertion points \\
213 & 8 & \textit{std} of composition score & 272 & 9 & \textit{std} of exertion points \\
214 & 8 & \textit{avg} revitalization score & 273 & 9 & \textit{avg} skin conductance levels (SCL) \\
215 & 8 & \textit{std} of revitalization score & 274 & 9 & \textit{std} of SCL \\
216 & 8 & \textit{avg} duration score & 275 & 9 & SCL variance \\
217 & 8 & \textit{std} of duration score & 276 & 9 & \textit{max} SCL \\
218 & 8 & \textit{avg} restlessness & 277 & 9 & \textit{min} SCL \\
219 & 8 & \textit{std} of restlessness & 278 & 9 & \begin{tabular}[c]{@{}c@{}}\textit{min}-\textit{max} difference\\ of SCL\end{tabular} \\
220-222 & 8 & \begin{tabular}[c]{@{}c@{}}\textit{avg} \{total, weekdays, weekend days\}\\ overall sleep score\end{tabular} & 279 & 9 & \textit{avg} \textit{std} of SCL \\
223 & 8 & \textit{std} of overall sleep score & 280 & 9 & \textit{std} of \textit{std} of SCL \\
224-226 & 8 & \begin{tabular}[c]{@{}c@{}}\textit{avg} \{total, weekdays, weekend days\}\\ efficiency of night sleeps\end{tabular} & 281 & 9 & \textit{std} of SCL variance \\
227-228 & 8 & \begin{tabular}[c]{@{}c@{}}\textit{avg} \{weekdays, weekend days\}\\ duration of night sleep (\textit{min}utes)\end{tabular} & 282 & 9 & \textit{max} \textit{std} of SCL \\
229-231 & 8 & \begin{tabular}[c]{@{}c@{}}\textit{avg} \{total, weekdays, weekend days\}\\ sleep start time\end{tabular} & 283 & 9 & \textit{min} \textit{std} of SCL \\
232-234 & 8 & \begin{tabular}[c]{@{}c@{}}\textit{avg} \{total, weekdays, weekend days\}\\ sleep end time\end{tabular} & 284 & 9 & \begin{tabular}[c]{@{}c@{}}\textit{min}-\textit{max} difference of \textit{std}\\ of SCL\end{tabular} \\
235-237 & 8 & \begin{tabular}[c]{@{}c@{}}\textit{avg} \{total, weekdays, weekend days\}\\ times waking up during night sleep\end{tabular} &  &  &  \\ \hline
\end{tabular}%
}
\end{table}

% \usepackage{graphicx}
\begin{table}[]
\caption{Results of evaluating the AI4FoodDB database using various state-of-the-art classifiers and feature selection methods. The performance is measured using the Area Under the Curve (AUC) metric. The best model for each dataset and feature selection method is highlighted in \textbf{bold}. SVM = Support Vector Machine, RF = Random Forest, LR = Logistic Regression, MLP = Multilayer Perceptron, GB = Gradient Boosting, XGB = eXtreme Gradient Boosting, KNN = K-Nearest Neighbors, SFFS = Sequential Forward Feature Selection.}
\label{tab:results_2}
\resizebox{\textwidth}{!}{%
\begin{tabular}{lcccc|cccc}
\cline{2-9}
\multicolumn{1}{c}{} & \textbf{SFFS} & \textbf{Boruta Selection} & \textbf{Genetic Algorithm} & \textbf{All Features} & \textbf{SFFS} & \textbf{Boruta Selection} & \textbf{Genetic Algorithm} & \textbf{All Features} \\ \cline{2-9} 
\textbf{} & \multicolumn{4}{c|}{\textbf{Dataset 4: Biomarkers}} & \multicolumn{4}{c}{\textbf{Dataset 6: Vital Signs}} \\ \hline
SVM & 47.96 & 64.24 & 47.00 & 48.24 & 42.82 & 45.32 & 44.02 & 43.66 \\
RF & 64.45 & 72.37 & 60.82 & 59.46 & 68.20 & 75.83 & 55.93 & 59.39 \\
LR & \textbf{69.86} & \textbf{74.69} & \textbf{62.00} & \textbf{67.45} & 52.39 & 58.33 & 52.46 & 51.84 \\
MLP & 50.70 & 74.15 & 61.93 & 51.87 & 52.00 & 63.31 & 53.15 & 45.24 \\
GB & 66.11 & 72.93 & 54.88 & 55.42 & 71.61 & \textbf{76.86} & \textbf{64.09} & \textbf{63.57} \\
{XGB} & {61.55} & {64.30} & {56.93} & {58.36} & {\textbf{73.47}} & {73.12} & {57.60} & {62.71}\\
KNN & 58.08 & 68.63 & 61.76 & 54.86 & 60.96 & 62.64 & 56.29 & 48.03 \\ \hline
\textbf{} & \multicolumn{4}{c|}{\textbf{Dataset 7: Physical Activity}} & \multicolumn{4}{c}{\textbf{Dataset 8: Sleep Activity}} \\ \hline
SVM & 43.71 & 50.71 & 45.65 & 46.07 & 47.61 & 58.51 & 42.84 & 42.21 \\
RF & \textbf{67.88} & \textbf{69.31} & 62.54 & 63.78 & \textbf{63.88} & 71.72 & \textbf{61.46} & \textbf{62.22} \\
LR & 55.41 & 61.24 & 57.86 & 55.02 & 54.36 & 60.77 & 56.26 & 54.83 \\
MLP & 48.70 & 51.95 & 55.85 & 45.63 & 52.06 & \textbf{72.00} & 47.53 & 48.18 \\
GB & 67.06 & 63.52 & \textbf{75.65} & \textbf{65.49} & 62.34 & 66.97 & 54.33 & 52.61 \\
{XGB} & {64.02} & {67.50} & {68.13} & {64.10} & {61.67} & {70.02} & {59.07} & {57.94}\\
KNN & 59.58 & 56.19 & 54.21 & 64.66 & 46.40 & 61.32 & 44.40 & 42.21 \\ \hline
\textbf{} & \multicolumn{4}{c|}{\textbf{Dataset 9: Emotional State}} & \multicolumn{4}{c}{\textbf{Combined Datasets}} \\ \hline
SVM & 49.39 & 51.64 & 47.34 & 47.36 & 54.48 & 66.47 & 57.48 & 52.92 \\
RF & 52.54 & 51.89 & 42.88 & 47.71 & 79.22 & 79.15 & \textbf{72.12} & \textbf{69.12} \\
LR & 48.72 & 52.86 & 50.13 & 48.18 & 65.88 & 72.06 & 71.05 & 67.72 \\
MLP & 49.32 & 54.28 & 32.61 & 51.12 & 62.72 & 69.48 & 63.68 & 57.60 \\
GB & \textbf{58.05} & 53.73 & \textbf{55.32} & \textbf{57.83} & \textbf{84.44} & \textbf{82.96} & 70.80 & 56.33 \\
{XGB} & {53.74} & {\textbf{57.84}} & {49.18} & {51.67} & {81.56} & {76.52} & {61.68} & {64.38}\\
KNN & 47.95 & 51.35 & 42.61 & 52.49 & 57.95 & 65.61 & 61.58 & 63.64 \\ \hline
\end{tabular}%
}
\end{table}

\end{document}